\documentclass[10pt,twocolumn,letterpaper]{article}

\usepackage{iccv}
\usepackage{times}
\usepackage{epsfig}
\usepackage{graphicx}
\usepackage{amsmath}
\usepackage{amssymb}
\usepackage{pgf}
\usepackage{siunitx}
\usepackage{etoolbox}
\usepackage{subcaption}
\usepackage{multirow}
\usepackage{booktabs}
\usepackage{textcomp}
\usepackage[final]{microtype}
\usepackage{paralist}
\usepackage{xspace}
\usepackage{placeins}

\usepackage[pagebackref=true,breaklinks=true,colorlinks,bookmarks=false]{hyperref}

\iccvfinalcopy 

\newcommand{\hide}[2]{\ificcvfinal#1\else#2\fi}

\def\iccvPaperID{4} 

\ificcvfinal\fi

\DeclareSIUnit{\nothing}{\relax}
\SendSettingsToPgf{}
\robustify\bfseries

\DeclareMathOperator*{\argmin}{argmin}

\newcommand\teq{\mkern0.0mu{=}\mkern1.5mu}
\newcommand\faze{\textsc{Faze}\xspace}
\newcommand\spaze{\textsc{Spaze}\xspace}

\begin{document}

\title{Learning to Personalize in Appearance-Based Gaze Tracking}

\author{\
	Erik Lind\'{e}n\\
	Tobii\\
	{\tt\small elin@tobii.com}
	\and
	Jonas Sj\"ostrand\\
	Tobii\\
	{\tt\small jsjd@tobii.com}
	\and
	Alexandre Proutiere\\
	KTH Royal Institute of Technology\\
	{\tt\small alepro@kth.se}
}

\maketitle
\ificcvfinal\thispagestyle{empty}\fi

\begin{abstract}
	Personal variations severely limit the performance of
	appearance-based gaze tracking. Adapting to these variations using standard
	neural network model adaptation methods is difficult. The problems range
	from overfitting, due to small amounts of training data, to underfitting,
	due to restrictive model architectures. We tackle these problems by
	introducing the SPatial Adaptive GaZe Estimator (\spaze). By modeling personal
	variations as a low-dimensional latent parameter space, \spaze provides just
	enough adaptability to capture the range of personal variations without
	being prone to overfitting. Calibrating \spaze for a new person reduces to
	solving a small optimization problem. \spaze achieves an error of
	\ang{2.70} with \num{9} calibration samples on MPIIGaze, improving on the
	state-of-the-art by \SI{14}{\percent}.

	We contribute to gaze tracking research by empirically showing that
	personal variations are well-modeled as a \num{3}\nobreakdash-dimensional
	latent parameter space for each eye. We show that this low-dimensionality
	is expected by examining model-based approaches to gaze tracking. We also
	show that accurate head pose-free gaze tracking is possible.

\end{abstract}

\section{Introduction}

Video-based gaze tracking deals with the problem of determining the gaze of a
person's eye given images of the eyes. By ``gaze'' one usually means the point
on a two-dimensional screen where the person is looking (2D gaze), but
sometimes one wishes to determine the complete gaze ray in 3D space,
originating from the eye and directed towards the screen gaze point. We refer
to this (five-dimensional) quantity as \emph{3D gaze}. Gaze tracking has
numerous applications: It is used as a communication aid for people with
medical disorders. Experimental psychologists use it to study human behavior.
In the consumer market, it is used for human-computer interaction and when used
in virtual reality, it can reduce the computational requirements through
foveated rendering, that is, rendering in high resolution only where the person
is looking.

Gaze tracking techniques can be categorized into model-based and
appearance-based methods~\cite{Hansen2010}. Model-based methods use image
features such as the pupil center and the iris edge, combined with a geometric
eye model to estimate the gaze direction. Some model-based methods also use the
corneal reflections from one or more light sources. These reflections are known
as \emph{glints} and the light sources are known as \emph{illuminators},
typically light-emitting diodes. Model-based methods can be implemented with
small amounts of training data, since they make simplifying assumptions. For
example, they might model the pupil as a dark ellipse. However, the same
assumptions make them unsuited to handle large variations in appearance.

Appearance-based methods, on the other hand, do not rely on hand-crafted
features. Instead, they estimate the gaze direction directly from the eye
images. This requires a larger amount of training data, but makes it possible
to track gazes despite appearance variations. Since appearance-based methods do
not require an explicit feature-extraction step, they are believed to work
better than model-based methods on low-resolution images~\cite{Zhang2015}.
Recent research on appearance-based methods using convolutional neural networks
\cite{Baluja1994, Tan2002, Krafka2016, Zhang2017, Zhang2017a, Zhang2018,
Deng2017} has focused on challenging in-the-wild scenarios with
person-independent models using low-resolution web camera images, and provided
promising results.

In this paper, we propose a new approach to personal calibration in
appearance-based gaze tracking: \spaze (SPatial Adaptive GaZe Estimator). We
analyze the performance of \spaze on both low-resolution webcam images and on
high-resolution images from near-infrared (NIR) cameras with active
illumination. \spaze consists of three steps:
\begin{inparaenum}[1)]
	\item From an image of the face, we extract three normalized images, two
	high-resolution images of the eyes and one low-resolution image of
	the face.
	\item The three images are fed into separate convolutional neural networks.
	\item The outputs of the networks are combined with
	person-specific calibration parameters in a fully connected layer. The
	calibration parameters are tuned by having the person look at known
	gaze targets.
\end{inparaenum}
Previous methods for personal calibration have trained person-specific models
\cite{Zhang2015,Zhang2017} or seen calibration as a post-processing step
\cite{Krafka2016}. This contrasts with \spaze where we include
calibration in the learning process, as a set of latent parameters for each
person.

We aim to estimate gaze rays in 3D space. To make data collection
simpler, we aim to learn this without ground truth for the 3D eye positions.
We only require ground truth for the 3D gaze targets.
By estimating 3D gaze and having a geometry-agnostic personal calibration,
we hope to learn a single model that can be used with different camera/screen
geometries.

\spaze is evaluated on three datasets:
\begin{inparaenum}[1)]
	\item a large \hide{Tobii}{} dataset from high-resolution, near-infrared
	cameras with active illumination,
	\item GazeCapture~\cite{Krafka2016} from iPhones and iPads and
	\item MPIIGaze~\cite{Zhang2017} from laptop webcams.
\end{inparaenum}

\noindent
Our experiments demonstrate that:
\begin{compactitem}
	\item \spaze is as accurate as model-based gaze tracking methods on
	high-resolution, near-infrared images.
	\item Our calibration achieves state-of-the-art performance on both
	within-MPIIGaze evaluations and on cross-dataset evaluations,
	transferring from GazeCapture to MPIIGaze.
	\item \spaze works without the help of a head pose estimator, using only a
	tight crop around the eyes.
	\item Personal variations are well-modeled as a
	\num{3}\nobreakdash-dimensional latent parameter space for each eye.
\end{compactitem}

The contributions and outline of the paper are as follows: In
Section~\ref{sec:relatedwork}, we review related work on appearance-based gaze
tracking and approaches to personal calibration. In Sections~\ref{sec:method}
and~\ref{sec:datasets}, we describe \spaze and the datasets we used. In
Section~\ref{sec:experiments}, we present experimental results. We find that
personal variations can be modeled as a \num{3}\nobreakdash-dimensional latent parameter
space for each eye. In Section~\ref{sec:modelbased}, we argue why this is
expected, reviewing typical model-based methods for gaze tracking. In
Section~\ref{sec:conclusions}, we discuss the implications of our results for
appearance-based gaze tracking.

\section{Related work}\label{sec:relatedwork}

Appearance-based gaze tracking has received much attention recently. Sugano
\etal~\cite{Sugano2014} used random forest regression to estimate gaze angles
from eye images. They introduced the normalization technique we adopt in this
paper, where the eye images are warped into a normalized camera view. This
effectively reduced the appearance variations their regressor had to handle.
For training, they augmented their dataset by rendering eye images from point
clouds. Zhang \etal~\cite{Zhang2015} introduced MPIIGaze, a dataset with
\SI{38}{\kilo\nothing} images from \num{15} persons. The dataset includes camera
calibration parameters and 3D gaze targets. They trained a light-weight
convolutional neural network to estimate gaze angles from eye images. Krafka
\etal~\cite{Krafka2016} collected a dataset of \SI{2.5}{\mega\nothing} images
taken with smartphones in uncontrolled environments. They trained a convolutional
neural network and without personal calibration they obtained an accuracy of
about \ang{3} (\SI{2}{\centi\meter}) on a phone or tablet. While the dataset is
large, it lacks camera calibration parameters and 3D ground truth for the gaze
targets. Wood \etal~\cite{Wood2016} introduced UnityEyes, a framework for
generating synthetic eye images which look realistic. With a $k$-nearest
neighbor regressor, they achieved an accuracy of \ang{10} on MPIIGaze.
Park \etal~\cite{Park2018} used a novel heat map approach to improve
the regression performance, achieving \ang{4.5} on MPIIGaze. Fischer
\etal~\cite{Fischer2018} trained an ensemble of 4 VGG networks, achieving
\ang{4.3} on MPIIGaze.

\subsection{Approaches to personal calibration}

In model-based gaze tracking there is often some kind of personal calibration
involved~\cite{Hansen2010}. The calibration parameters typically include the fovea
offset for each eye, as discussed in Section~\ref{sec:modelbased}. In
appearance-based methods there is no explicit model of the eye, so it is not
clear how to incorporate personal calibration. Most works ignore calibration,
though some authors have shown that person-specific estimators greatly
outperform generic estimators.

For example, Sugano \etal~\cite{Sugano2014} compared random forest regression
with person-specific training and cross-person training,
finding errors of \ang{3.9} and \ang{6.5}. Zhang \etal~\cite{Zhang2015} made
the same comparison using a convolutional neural network on a different
dataset, finding \ang{3.3} and \ang{6.3}. The comparison on MPIIGaze found
\ang{2.5} and \ang{5.4}~\cite{Zhang2017}. Building on the idea of training
person-specific estimators, Zhang \etal~\cite{Zhang2018} devised a method which
helped to collect more training data for a specific person.

A neural network trained for a specific person will of course always outperform
a generic network, assuming equal amounts of training data. However, it is
unpractical to collect the vast amount of data needed to train a modern neural
network from a single person.

One approach to this problem is to fine-tune some part of a generic network,
typically the final layer. We adopt this
approach as in iTracker by Krafka \etal~\cite{Krafka2016}, but with a critical
difference. In iTracker, the network is initially trained to learn an
average model for all persons present in the dataset. The calibration is just
a post-processing step where the last fully connected layer with input size \num{128}
and output size \num{2} is replaced by an SVR-model trained with calibration samples
for a specific person (while the weights of the rest of the network are kept
fixed). With \num{13} calibration points, the accuracy was improved with up to
\SI{20}{\percent}, whereas with only \num{4} calibration points, the accuracy
was actually worse than without personal calibration, probably due to
overfitting of the SVR-model. In iTracker, personal calibration is
learnt from scratch for each person. This requires a large calibration
parameter space, to support a sufficiently rich set of calibration mappings,
and hence a large number of calibration samples. This contrasts with our
method: We parameterize the set of possible calibration mappings by a neural
network, and during the initial training, the network learns a suitable mapping
from a small calibration parameter space. The high expressiveness of the neural
network means it can potentially model many different calibration mappings, and
the training must find a suitable one. This necessitates a large training set,
to prevent overfitting. We have this large dataset, since we learn the
calibration mapping from the whole training set, not just from one person.

Park \etal~\cite{Park2018a}
adapted the method of iTracker by extracting \num{32}
landmarks (corresponding to iris edges, pupils, eyelids). These features are
then used to train a personal SVR as in iTracker.

Liu \etal~\cite{Liu2018}
implemented personal calibration by training a differential neural network to
estimate the difference in gaze direction between two images.
When estimating the gaze direction in a novel image, they used a set of
calibration images as anchors, estimating the gaze direction in the
novel image relative the calibration images.
This achieved \ang{4.67} with \num{9} calibration samples on MPIIGaze.
In an inversion of this approach, Yu \etal~\cite{ImprovingFewShot}
used gaze redirection to synthesize additional person-specific training
images. This extended calibration set was then used to fine-tune their network,
achieving \ang{4.01} with \num{9} calibration samples on
MPIIGaze.

The state-of-the-art method for calibrated gaze tracking is \faze (Few-shot
Adaptive GaZE Estimation) by Park \etal~\cite{FewshotAdaptive}. \faze uses an
encoder/decoder structure to learn a compact and consistent gaze
representation. This gaze representation is sent to a small multi-layer
perceptron (\num{64} hidden units). The perceptron is trained using model-agnostic
meta-learning (MAML) to allow fast person-specific fine-tuning with minimal
overfitting. \faze achieved \ang{3.14} with \num{9} calibration samples on
MPIIGaze, when trained on GazeCapture. The performance was reduced when \faze was
trained on the much smaller MPIIGaze.

\section{The \spaze method}\label{sec:method}

\begin{figure}
	\centering
	\includegraphics{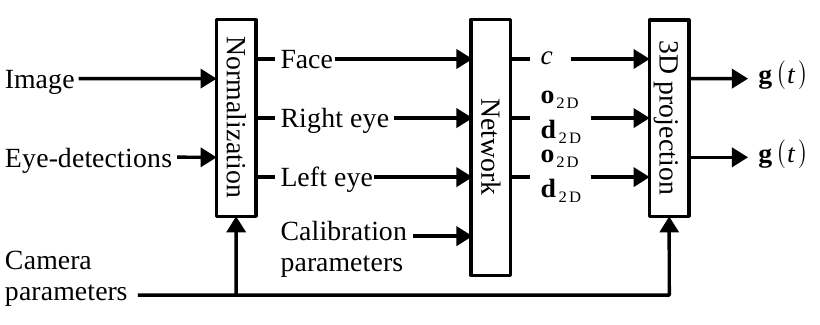}
	\caption{\
		Network pre- and post-processing for image normalization and 3D
		gaze projection.}\label{fig:net_overview}
	\vspace{-1mm}
\end{figure}

\begin{figure*}
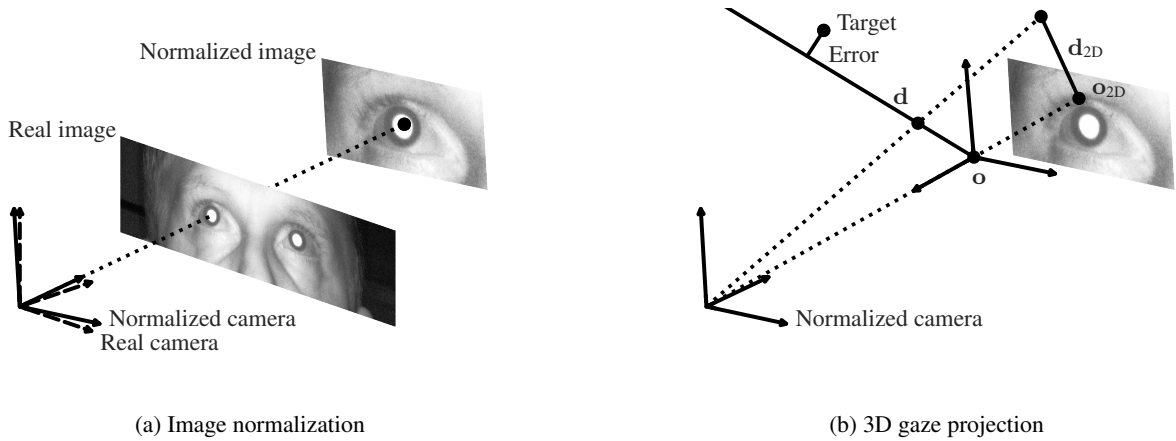

	\centering
	\begin{subfigure}{\columnwidth}
		\centering
		\input{imgs/geo1.pgf}
		\caption{Image normalization}\label{fig:geo1}
	\end{subfigure}
	\hfill
	\begin{subfigure}{\columnwidth}
		\centering
		\input{imgs/geo2.pgf}
		\caption{3D gaze projection}\label{fig:geo2}
	\end{subfigure}
	\caption{\
		(a) The image captured by the physical camera is warped into a
		normalized camera looking directly at the reference point, in this case
		the persons right eye. (b) The 2D gaze origin and gaze direction are
		combined with the corrected distance to form a gaze ray in 3D space.
		The miss distance between the gaze ray and the gaze target is the loss
	used to train the neural network.}
	\vspace{-1mm}
\end{figure*}

In this section, we describe the three main components of \spaze:
\begin{inparaenum}[1)]
	\item image normalization,
	\item a neural network and
	\item 3D gaze projection.
\end{inparaenum}
See
Figure~\ref{fig:net_overview} for an overview of the data flow. We also
describe our personal calibration method.

\subsection{Image normalization}

To improve generalization, we normalize the images as described by
\cite{Zhang2018a}. Unlike most other methods, we do not use a separate head
pose estimator for image normalization and 3D gaze estimation. We have found
head pose estimators to be fragile and they require pose-annotated training
data. Our NIR datasets is also so tightly cropped to the eyes that most of the
head is not visible. Instead we adopt a structure where eye detections are used
for a rough pose estimation, for image normalization. The network is then asked
to correct any errors (affecting gaze accuracy) in the rough estimate. The idea
of head pose-free methods has previously been investigated by~\cite{Lu2011,
Deng2017}.

The input to the image normalization component is an image of a person's face
and two points in the image defining where the eyes are. Those points are
provided by an external eye detector. The output is three images: two
high-resolution eye images centered at the eye detection points and one
low-resolution face image centered at the midpoint between the eyes.

By assuming that depth variations in the face are small compared to the distance
between the face and the camera, we can compensate for arbitrary scaling and
camera rotation by a perspective image warp. This reduces the complexity of the
gaze tracking problem, as the estimator does not need to handle arbitrary
face rotations or scalings. However, due to imperfections in the normalization
method, some rotation and scaling errors will remain.

Figure~\ref{fig:geo1} illustrates the normalization. Given an input image
$\mathbf{I}$ and a reference point (either an eye detection point or the
midpoint between the eyes), we compute a conversion matrix
$\mathbf{R}$.
Its
inverse $\mathbf{R}^{-1}$ is the matrix that rotates the camera so that it
looks at the reference point and so that the interocular vector in the image
becomes parallel to the camera $x$-axis. To make the eye appearance consistent,
for the left-eye image we also let $\mathbf{R}^{-1}$ mirror the camera in the
interocular direction after the rotation.

The conversion matrix $\mathbf{R}$ will map any 3D point in the real camera
coordinate system into the normalized camera coordinate system. The same
transform is applied to the image $\mathbf{I}$ using an image transformation
matrix $\mathbf{C}_n \mathbf{R} \mathbf{C}_r^{-1}$, where
$\mathbf{C}_r$ is the projection matrix of the real camera and $\mathbf{C}_n$
is the projection matrix of the normalized camera. $\mathbf{C}_n$ is selected
as a scaling such that the interocular distance in the normalized image becomes
\num{320}~pixels for the eye images and \num{84}~pixels for the face image. We
use bilinear interpolation to implement the warping and crop out a $W \times H$
region in the normalized image, \num{224 x 112}~pixels for the eye images and
\num{224 x 56}~pixels for the face image.

A gaze ray $\widehat{\mathbf{g}}(t) = \mathbf{o} + t\mathbf{d}$ is estimated in
the normalized camera coordinate system and transformed back to the real camera
coordinate system by $\mathbf{g}(t) = \mathbf{R}^{-1}\widehat{\mathbf{g}}(t)$.

\subsection{3D gaze projection}

Here we describe how the output from the network is translated into a pair of
gaze rays. The network has five outputs. For each eye, it predicts a 2D gaze
origin $\mathbf{o}_\text{2D}$ and a 2D gaze direction $\mathbf{d}_\text{2D}$.
It also generates a distance correction term, $c$, which is common to both eyes.
We assume that the distance from eye to camera is approximately the same for
both eyes, and our estimate of that distance will be called $\rho$. First, given the input
image and the eye detections, we find a rough distance $\rho_\text{rough}$ such
that the separation between the eyes becomes \SI{63}{\milli\meter} at that
distance, approximately the average human interocular distance
\cite{Gordon2014}. This distance is then corrected by the network by setting
$\rho = c \, \rho_\text{rough}$. The rough distance will be unreliable, since
it is based on only the eye detections, which are noisy. Further, it makes no
allowance for head yaw. But since the same eye detections are used to normalize
the images fed to the network, the network has an opportunity to spot
misaligned eye detections and correct for them. Likewise, it can measure the
head yaw and correct for it.

We will now describe how a gaze ray is computed for a single eye, see
Figure~\ref{fig:geo2} for an overview of the procedure. The 3D origin of the
gaze ray, $\mathbf{o}$, is computed by back-projecting the 2D gaze origin
$\mathbf{o}_\text{2D}$ through the normalized camera to the distance $\rho$. To
compute the 3D direction of the gaze ray, $\mathbf{d}$, we first construct a
set of orthonormal basis vectors $\{\mathbf{x},\mathbf{y},\mathbf{z}\}$, where
$\mathbf{z}$ points from the 3D gaze origin to the camera and $\mathbf{x}$ is
orthogonal to the $y$-axis of the normalized camera coordinate system. The 3D
gaze direction $\mathbf{d}$ is then computed from the 2D gaze direction
$\mathbf{d}_\text{2D}$ as
$\mathbf{d} = \left[\mathbf{x} \, \mathbf{y}\right] \mathbf{d}_\text{2D} + \mathbf{z}$.

\subsection{Network architecture}

\begin{figure}
	\centering
	\includegraphics{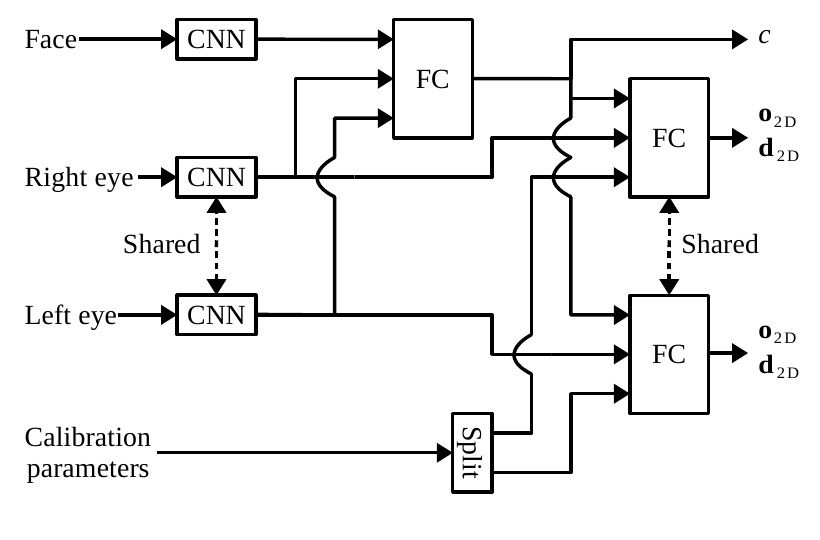}
	\caption{Network architecture.}\label{fig:net_arch}
	\vspace{-1mm}
\end{figure}

Here we describe the input to the network and how the output is generated.
See Figure~\ref{fig:net_arch} for an overview of the network
architecture. We feed three images to the network: both eyes at high resolution
and the face at low resolution. One convolutional network is applied separately
to each eye. Another convolutional network is applied to the face. Both
networks are the convolutional part of ResNet-18~\cite{He2015}.

The three outputs are concatenated and fed to a fully connected module. The
fully connected module outputs the distance correction $c$. The rationale for
using both the eyes and the face for the distance estimation is that the eyes
could provide accurate interocular distance measurements, while the face could
improve head pose estimation.

The convolutional network output for each eye is then concatenated with a set
of $N$ personal calibration parameters and the distance correction. This
combined feature vector is fed to a fully connected module. The fully connected
module outputs the 2D gaze origin $\mathbf{o}_\text{2D}$ and the 2D gaze
direction $\mathbf{d}_\text{2D}$. The same module is used for both eyes.

\newcommand{\nnr}{{\rightarrow}\allowbreak}

The fully connected modules can both be described as:
$\mathrm{FC}(3072)\nnr\mathrm{BN}\nnr\mathrm{ReLU}\nnr\mathrm{DO}\nnr
\mathrm{FC}(3072)\nnr\mathrm{BN}\nnr\mathrm{ReLU}\nnr\mathrm{DO}\nnr
\mathrm{FC}(1\text{ or }4)$ where $\mathrm{FC}$ is fully connected,
$\mathrm{BN}$ is batch normalization and $\mathrm{DO}$ is dropout.

Initially we generated both gaze rays using information from
both eyes. That gave better performance, but made the gaze rays for the two
eyes highly correlated, as all training data have both eyes looking at the same
point. As we want to support medical applications, where it is important to
observe differences between the eyes, we choose this segregated approach. In a
consumer application, it would be more appropriate to modify the network
architectures to estimate a joint gaze direction, originating from between the
eyes, as suggested by Zhang \etal~\cite{Zhang2017a}.

The rationale for providing the distance correction $c$ when estimating the
gaze direction is to allow the use of features that require accurate distance
information. This is the case with pupil-center/corneal-reflection gaze mapping
\cite{Hansen2010, Guestrin2006}. We do not input personal calibration
parameters into the distance-estimation module, since it is typically impossible
to detect distance errors from calibration data collected at a single distance,
which is what we have.

\subsection{Training and calibration}

Personal variations are modeled by assigning $2N$ calibration parameters to
each person, $N$ for each eye. During training, the training set $\mathcal{D}$
consists of triples $(X, t, m)$, where $X$ is a triple of images (face, right
eye, left eye), $t$ is the 3D gaze target and $m$ is the index of the person in
the images. With $M$ persons in the training set, we solve the optimization
problem\footnote{Note that our objective function includes person-specific
parameters $P_m$, which contrasts with existing methods~\cite{Krafka2016} whose
objective function does not account for personal variations (with our
notations, it would be
$\sum_{(X,t)\in\mathcal{D}}\mathrm{loss}(g_\theta(X),t)$).}
\[
	(\theta_{\rm opt}, P_{\rm opt})=\argmin_{\theta, P}\sum_{(X,t,m)\in\mathcal{D}}\mathrm{loss}(g_\theta(X, P_{m,\cdot}), t)
\]
over all weights $\theta$ of the neural network $g$ and, simultaneously,
over all $M$-by-$2N$ matrices $P$ of calibration parameters. After
training, $P_{\rm opt}$ is discarded. The optimization method and the loss
function will be described in a moment.

To calibrate for a new person, we collect a small calibration set
$\mathcal{D}_\mathrm{cal}$ of (image triple, gaze target) pairs. The
calibration procedure then finds the person's $2N$ calibration parameter vector
defined as
\[
	\argmin_{p\in\mathbb{R}^{2N}}\sum_{(X,t)\in\mathcal{D}_\mathrm{cal}}\mathrm{loss}(g_{\theta_{\rm opt}}(X, p), t).
\]
Note that the network weights are fixed.
We used BFGS to solve this optimization problem,
and as an initial guess we find a single calibration parameter vector
that minimizes the error over the whole
training set. This parameter vector is also used for
the uncalibrated case.

In the experiments, we vary $N$ and find that \num{3} parameters per eye are
enough to provide an efficient person-specific gaze tracking. Modeling
personal variations using such a low-dimensional latent parameter space
contrasts with existing calibration approaches (e.g.\ in iTracker, a
much higher number of parameters is used), but can be motivated and justified
by existing model-based methods; see Section~\ref{sec:modelbased}
for a complete discussion.

Finally, let us describe the loss function and the optimization method during
training. The loss is the miss distance between the gaze ray and the 3D gaze
target, see Figure~\ref{fig:geo2}. We train using Adam~\cite{Kingma2014} with a
learning rate of \num{e-3} and a weight decay of \num{e-5}. We train for
\num{30} epochs, decaying the learning rate by \num{10} every \num{10}th epoch.
The eye detections are jittered for data augmentation in training: We randomly
offset the detections in a disk with a radius equal to \SI{4}{\percent} of the
interocular distance.

With this setup, we found that the generated gaze rays would pass close to the
gaze target, but the 3D gaze origin would be highly incorrect. The reason is
that current datasets are collected with a geometry that makes the 3D
gaze origin underdetermined. In the supplementary material, we use a synthetic
dataset to investigate whether this limitation can be overcome by a different
data collection strategy.

To prevent unphysical solutions when training on current datasets, we
introduced two regularizing terms:
\begin{inparaenum}[1)]
	\item a hinge loss on the 2D gaze origin
	$\mathbf{o}_\text{2D}$, penalizing if it moves outside the eye image and
	\item a hinge loss on the distance correction $c$, penalizing changes in
	distance by more than~\SI{40}{\percent} in either direction.
\end{inparaenum}

\section{Datasets}\label{sec:datasets}

We use three datasets in our experiments:
\begin{inparaenum}[1)]
	\item a \hide{Tobii}{} dataset from high-resolution, near-infrared (NIR) cameras,
	\item GazeCapture and
	\item MPIIGaze.
\end{inparaenum}
In the supplementary material we investigate the effect of gaze
target placement and glints using synthetic UnityEyes images.

\subsection{NIR dataset}

We use a large \hide{Tobii}{} dataset for our calibration
experiments. The dataset was collected with a near-infrared (NIR) eye tracker
platform, with an infrared illuminator mounted very close to the camera. This
produces a bright-pupil effect~\cite{Hansen2010}, the same effect that makes
the eyes red in flash photography. Since the illuminator position coincides
with the camera position, we can scale and rotate the normalized camera without
changing the position of the illuminator in the camera coordinate system. The
training set was collected over a period of several years and contains
\SI{427}{\kilo\nothing}~images from \num{1824}~persons. The majority of the training data
have gaze targets on a regular lattice on the screen.

For evaluation, we have a set of \num{200}~persons. This set was
collected on \num{19}~inch, \num{16}:\num{10}~aspect ratio screens, with the
camera mounted at the bottom edge of the screen and tilted up \ang{20}. The
camera focal length was \num{3679}~pixels and it captured a region of interest
of \num{1150 x 300}~pixels. The region of interest was kept aligned on the eyes
using an eye detector. The persons sat at \SI{65(10)}{\centi\meter} from the
camera. See Figure~\ref{fig:geo1} for an example image. Half of the recordings
were made in \hide{Sweden}{Europe}, the other half in \hide{China}{Asia}.

There are three recordings for each person, one for calibration and two for
evaluation. From each recording, we extract \num{45} images, evenly distributed over
the gaze targets. The calibration recordings have the gaze targets on a regular
\num{3 x 3}~lattice. For the evaluation recordings, the screen was divided into a
\num{3 x 4}~grid and a gaze target was placed randomly in each grid cell. The screen
brightness was also randomized. For evaluation, we first calibrate
on the calibration recording and then report the error on the two evaluation
recordings. As the setup was controlled, we believe head yaw angles were
on the order of \ang{10}.

\subsection{RGB datasets}

\noindent
\textbf{GazeCapture}~\cite{Krafka2016} contains \SI{1.5}{\mega\nothing} images
from \num{1471} persons, collected on iPhones and iPads. It does not provide
camera calibrations. To apply \spaze, we assume that all cameras have a
horizontal field-of-view of \ang{54.4}~\cite{Apple} and that the principal
point is in the middle of the image. As the eye detections are poor, we
increase the area viewed by our eye images by \SI{50}{\percent}. For evaluation, we
use those \num{105} persons in the test set having at least \num{756} images. We
report the same error metric as Krafka \etal, the on-screen miss distance,
since GazeCapture does not provide the 3D data needed to compute angles.

\noindent
\textbf{MPIIGaze}~\cite{Zhang2018} is a widely used benchmark for gaze
estimation. It consists of webcam images from \num{15} persons. We used the
MPIIFaceGaze~\cite{Zhang2017a} subset, which has around \num{2500} images per
person.

We adopt the experimental protocol of \faze~\cite{FewshotAdaptive} where the
last \num{500} images for each person are used for evaluation and $k$
calibration samples are drawn randomly from the remaining images. For each $k$,
we adaptively control the total number of trials to keep the uncertainty in the
final error below one-hundredth of a degree.\footnote{The mean error is based on about
\SI{13}{\kilo\nothing} individual calibrations for $k{=}1$, down to \num{150}
calibrations for $k{=}256$.} For cross-dataset evaluation, we train on the
training set of GazeCapture and evaluate on MPIIGaze. For within-dataset
evaluation, we perform leave-one-out training on MPIIGaze.

\section{Experiments}\label{sec:experiments}

While we minimize the gaze ray miss distance, we report the error as an angle
for easier comparison with previous works. Since we estimate a 3D gaze ray, we
compute the angle as follows: Let $\mathbf{ET}$ be the line through the true
(annotated) 3D eye center and the 3D gaze target. Then we find the point
$\mathbf{G}$ on the gaze ray $\mathbf{g}(t)$ where the line
$\mathbf{GT}$ is orthogonal to $\mathbf{ET}$. We measure the error as the
angle between $\mathbf{ET}$ and $\mathbf{EG}$.

In some figures, we show shaded bands around the mean value. These show the
\emph{consistency} of the calibration, how much a user can expect the
mean error to vary between repeated calibrations.\footnote{The one-sided band
width is $\sqrt{\frac{1}{M}\sum_{m=1}^M\mathrm{Var}(\mu_m)}$ where
${\{\mu_m\}}_{m=1}^M$ is the random variable of the mean error for each person.
The randomness comes from the random selection of calibration samples.}

\subsection{Number of calibration parameters}

\begin{figure*}
	\centering
	\begin{subfigure}{\columnwidth}
		\centering
		\input{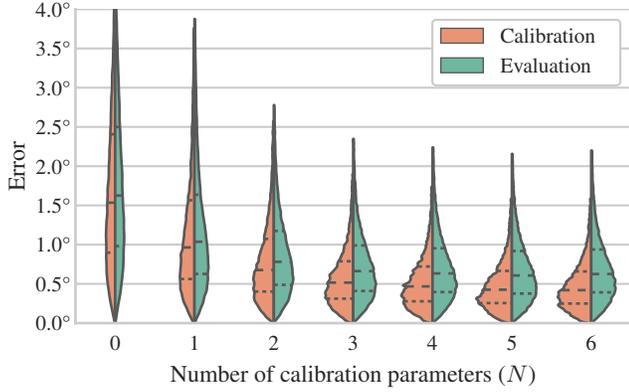}
		\caption{NIR dataset}\label{fig:n_params}
	\end{subfigure}
	\hfill
	\begin{subfigure}{\columnwidth}
		\centering
		\input{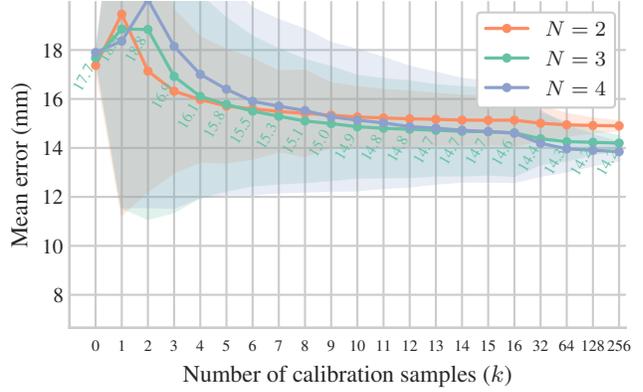}
		\caption{GazeCapture (test)}\label{fig:n_params_gc}
	\end{subfigure}
	\begin{subfigure}{\columnwidth}
		\centering
		\input{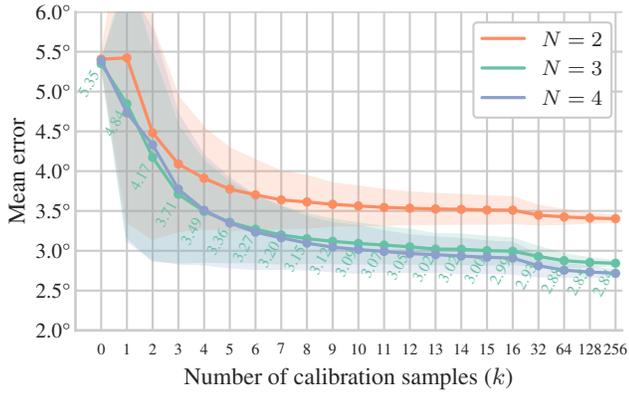}
		\caption{GazeCapture-to-MPIIGaze}\label{fig:n_params_gc_mpii}
	\end{subfigure}
	\hfill
	\begin{subfigure}{\columnwidth}
		\centering
		\input{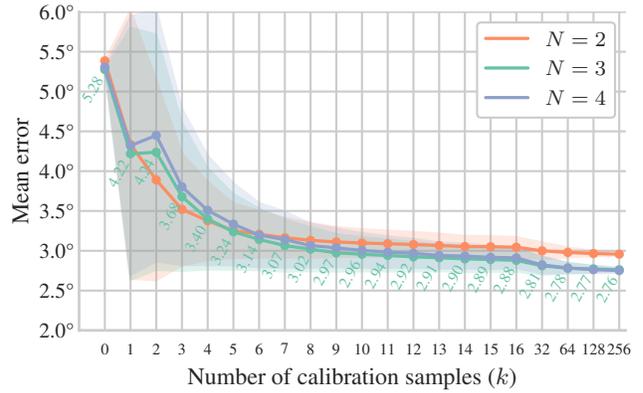}
		\caption{Within-MPIIGaze}\label{fig:n_params_mpii}
	\end{subfigure}
	\caption{Comparison of different numbers of calibration parameters ($N$).\label{fig:n_params_all}}
	\vspace{-1mm}
\end{figure*}

We vary the number of calibration parameters ($N$), see
Figure~\ref{fig:n_params_all}. In all cases, there is little improvement beyond
\num{3} parameters. We use $N\teq3$ in all subsequent experiments.

For the NIR dataset, (\hyperref[fig:n_params]{a}), we found that calibration
shifts the whole error distribution downwards. Specifically, any error quantile
falls to two-fifths of its previous value. This suggests that personal variations are
a dominating factor throughout the error distribution. Our mean
error with calibration is \ang{0.79} (uncalibrated \ang{1.85}), which
compares well with multi-camera, multi-illuminator systems~\cite{Hansen2010,
Ferhat2016}.

In (\hyperref[fig:n_params_gc]{b}) and (\hyperref[fig:n_params_mpii]{d}), we
see that fewer parameters are generally better for few calibration samples,
which is expected since the low dimensionality provides regularization. The
improvement from $N\teq2$ to $N\teq3$ is much larger for
GazeCapture-to-MPIIGaze, (\hyperref[fig:n_params_gc_mpii]{c}), than for
within-dataset evaluation, (\hyperref[fig:n_params_gc]{b}) and
(\hyperref[fig:n_params_mpii]{d}). Looking closer at the GazeCapture results,
we found that changing $N\teq2$ to $N\teq3$ mostly affected the error for
tablets ($\SI{21.0}{\milli\meter} \rightarrow \SI{17.0}{\milli\meter}$), while
phones remained similar ($\SI{12.5}{\milli\meter} \rightarrow
\SI{12.3}{\milli\meter}$). Tablets generate larger gaze angles than phones, but
tablets are a minority in GazeCapture. We suspect that with $N\teq2$, the
network is forced to choose between offset and scaling (see
Section~\ref{sec:modelbased}). For small gaze angles, like on phones, offset
might be more important than scaling. But for MPIIGaze, which has larger gaze
angles, scaling might be more important than offset.

\subsection{Comparison with state-of-the-art}

\begin{figure}
	\centering
	\input{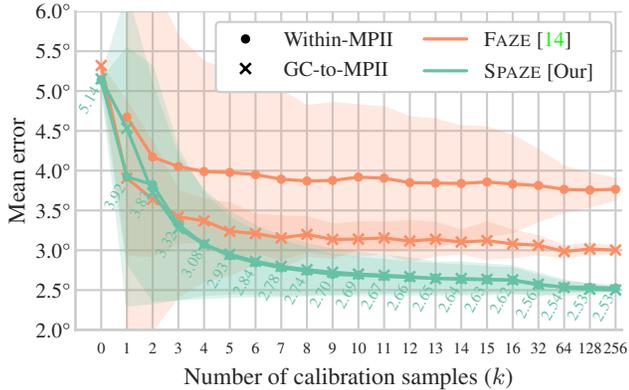}
	\caption{MPIIGaze. Comparison with \faze.}\label{fig:faze_comparison}
	\vspace{-1mm}
\end{figure}

\begin{table}
	\centering
	\setlength{\tabcolsep}{4.5pt}
	\begin{tabular}
		{
		l
		S[table-format=4]
		S[table-format=1.2]
		S[table-format=1.2,separate-uncertainty,table-figures-uncertainty=1]
		}
		\toprule
		\multicolumn{1}{l}{Method} &
		\multicolumn{1}{c}{Samples ($k$)} &
		\multicolumn{1}{c}{Their (\si{\degree})} &
		\multicolumn{1}{c}{Our (\si{\degree})} \\
		\midrule
		Diff-NN~\cite{Liu2018}                             & 9    & 4.67 & 2.94\pm0.22 \\
		\midrule
		GazeML~\cite{Park2018a}                            & 20   & 4.6  & 2.82\pm0.09 \\
		\midrule
		\multirow{3}{*}{RedFTAdap~\cite{ImprovingFewShot}} & 1    & 4.97 & 4.12\pm1.48 \\
		                                                   & 5    & 4.20 & 3.16\pm0.42 \\
		                                                   & 9    & 4.01 & 2.94\pm0.22 \\
		\midrule
		GazeNet+~\cite{Zhang2017}                          & 2500 & 2.5  & 2.76\pm0.00 \\
		\bottomrule
	\end{tabular}
	\caption{Comparison of \spaze with other calibrated methods on MPIIGaze.
		\spaze is evaluated using the experimental protocol of the compared method.}\label{tab:mpiigaze}
	\vspace{-1mm}
\end{table}

We compare \spaze with \faze~\cite{FewshotAdaptive}, a state-of-the-art method
for calibrated appearance-based gaze tracking. To closer match the \faze
experimental protocol, we form a single gaze ray by averaging the origins and
directions of the two individual gaze rays generated by our network.

See Figure~\ref{fig:faze_comparison} for the comparison with \faze.
Within-MPIIGaze training does not work well with \faze, while \spaze performs
equally well within MPIIGaze and transferred from GazeCapture to MPIIGaze.
Looking only at the \faze GazeCapture-to-MPIIGaze results, the two methods have
similar uncalibrated performance, but \spaze has a better mean error for
$k>3$. For $k\teq9$, we improve from \ang{3.14} to \ang{2.70}. \faze has better
consistency for $k<12$. \spaze does not consider the a priori
likelihood of the calibration parameters, it drives the empirical
calibration error as low as possible. This results in large variations for
small $k$.

We also compare \spaze with some other calibrated methods, see
Table~\ref{tab:mpiigaze}. \spaze outperforms the other methods, approaching
the performance of GazeNet+, which trains a person-specific network on
\num{2500} images. Another advantage of \spaze over most other methods is
the absence of hyper-parameters; there is no learning rate or early stopping to
tune. The calibration optimization uses the standard BFGS optimizer, operating
on \num{2 x 3} parameters, and drives the calibration error to a local minima.
Since only the fully connected layers depend on the calibration parameter, the
calibration optimization does not need to evaluate the expensive convolutional
layers, resulting in a fast calibration.

\subsection{Distance correction}

\begin{figure}
	\centering
	\input{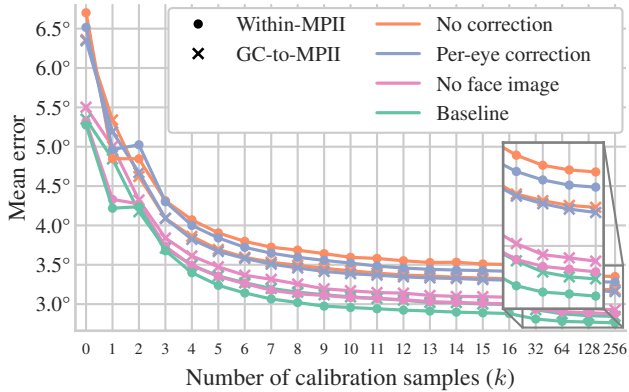}
	\caption{MPIIGaze. Different types of distance correction.}\label{fig:dist_corr}
	\vspace{-1mm}
\end{figure}

To validate our use of a distance correction derived from the two eye images and
the face image, we perform an ablation study where we:
\begin{inparaenum}[1)]
	\item disable the distance correction and only use the rough
	distance from the eye detections,
	\item use a per-eye distance correction, derived from only the image of that
	eye and
	\item derive the distance correction from only the two eye images, without the face image.
\end{inparaenum}

The results in Figure~\ref{fig:dist_corr} show that our method is superior to
the alternatives. The difference in performance with and without the distance
correction is much smaller with calibration (\ang{0.59}) than without
calibration (\ang{1.43}). This can be explained as follows: Even if the network
cannot explicitly correct the distance, it can compensate for an erroneous
distance by changing the gaze angle (see the supplementary material). This is
why the per-eye distance correction performs similarly to no distance
correction. Estimating the distance error from a single eye image is more
difficult than estimating the distance error from both eye images and the face
image. Calibration can alleviate this difficulty by providing person-specific
information on how to estimate the distance error from the single eye image.

\section{Low-dimensional parameter space for calibration: the case of model-based methods}\label{sec:modelbased}

To understand why it is reasonable to model personal variations as a
low-dimensional latent parameter space, it helps to look at a typical
model-based method for gaze tracking. Here we will review the eye model
described by Guestrin and Eizenman~\cite{Guestrin2006}. For a comprehensive
review of model-based methods, we refer to~\cite{Hansen2010}.

We will describe a gaze mapping model called pupil-center/corneal-reflection
(PCCR). Assume a system with one camera and a collocated illuminator.
Further assume we know the distance to the eye, from head pose estimation, a
second stereo camera or some other method.

Image processing methods detect the corneal reflection, the glint, from the
illuminator. The center of the pupil is also detected. The difference between
these two points forms the pupil-center/corneal-reflection vector. If the
cornea is assumed to be spherical, the cornea center lies directly behind the
glint. The \emph{optical axis}, a line passing through the cornea center and
the pupil center, can then be computed if the distance between the person's
cornea center and pupil center is known. This distance is one calibration
parameter.

However, the optical axis is not the \emph{visual axis}, the person's line of
gaze. The fovea, the most sensitive part of the retina, is offset from the
optical axis, and this offset, in two dimensions, differs from person to
person.

Taken together, we have three parameters per eye for each person. The foveal
offset roughly corresponds to shifting the gaze up-and-down and side-to-side,
and the cornea-center/pupil-center distance scales the gaze around the optical
axis.

Over the three datasets, NIR, GazeCapture and MPIIGaze, GazeCapture
sees the least improvement from calibration, and the NIR dataset sees the
greatest. The former also has the worst images, while the latter has the best.
We do not believe that this is a coincidence. As measurement errors in the
images decrease, the mismatch between a generic gaze tracking model and a
specific person's eye geometry becomes the dominant source of errors.

\section{Conclusions}\label{sec:conclusions}

We have presented \spaze, a method for calibrated appearance-based gaze
tracking. \spaze achieves state-of-the-art results on MPIIGaze and is as
accurate as model-based gaze tracking on high-resolution, near-infrared image.
Our results show that personal variations are well-modeled as a
\num{3}\nobreakdash-dimensional latent parameter space for each eye. Our
results also show that accurate gaze tracking is possible without a separate
head pose estimator.

\hide{\
	\paragraph{Acknowledgement}
	This work was partially supported by the Wallenberg AI, Autonomous Systems and
	Software Program (WASP) funded by the Knut and Alice Wallenberg Foundation
	}{}

\newpage
{\small
	\bibliographystyle{ieee}
	\bibliography{egbib}
}

\newpage
\section*{\Large Supplementary material}

\section*{On 3D gaze origins}

\begin{figure}
	\centering
	\includegraphics{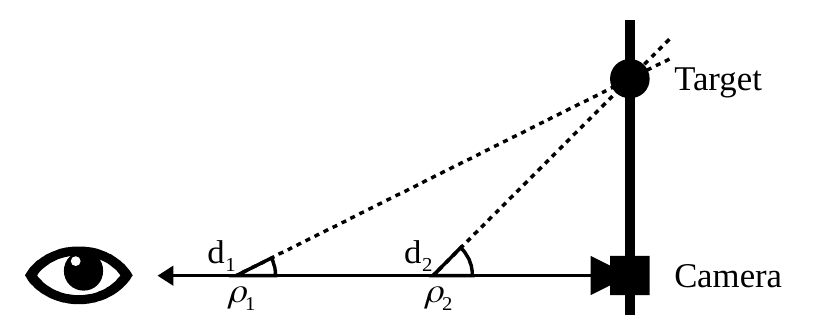}
	\caption{\
		When the eyes lie along the $z$-axis of the camera and the gaze targets
		are in the image plane of the camera, there are multiple combinations
		of distance and gaze direction that minimize the error.}\label{fig:distance_angle}
\end{figure}

The NIR dataset, GazeCapture and MPIIGaze all have the gaze targets in a
single plane (approximately the camera image plane). We found that this makes
it impossible to learn meaningful gaze origins. The network can always
compensate for an incorrect origin by modifying the gaze direction, see
Figure~\ref{fig:distance_angle}. The gaze ray will still pass through the gaze
target, but the ray is only correct at that point. However, if gaze targets are
placed at various $z$-depths in the camera coordinate system, it is no longer
possible to compensate origin errors by changing the gaze direction. The reason
is that there will be pairs of well separated gaze targets corresponding to
nearly identical input images to the network. By continuity, the corresponding
pair of gaze ray estimates are almost identical, and the error is minimized
only when that generated gaze ray almost passes through both gaze targets.

\FloatBarrier{}

The gaze origins could be learned directly if ground truth 3D eye positions are
available, but in practice we find it difficult to design a data collection
setup where we can measure the position of eyes with the necessary accuracy,
which is on the order of millimeters.

\subsection*{Targets in multiple planes}

\begin{figure}
	\centering
	\includegraphics[width=\linewidth]{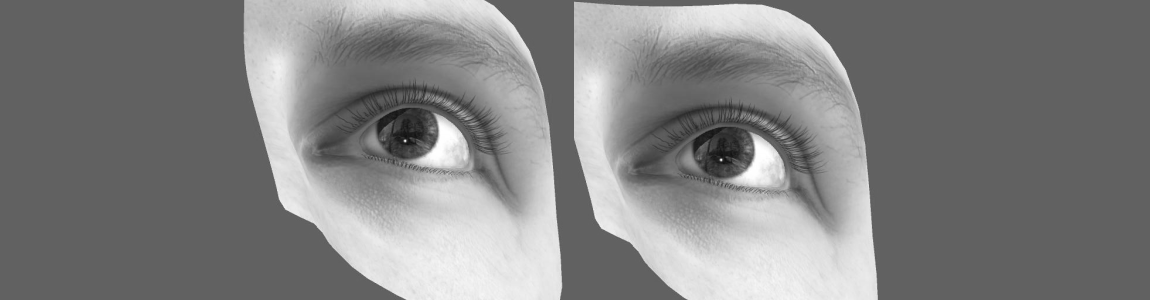}
	\caption{UnityEyes composition.}\label{fig:unity_roi}
\end{figure}

To test the feasibility of learning 3D gaze without ground-truth eye positions,
we generate synthetic images using the UnityEyes tool~\cite{Wood2016}. This
lets us place gaze targets at different depths. The dataset defines gaze in
terms of the \emph{optical axis} of the eye~\cite{Hansen2010}, so we use no
personal calibration.

We make the synthetic images similar to the NIR dataset, with the camera
\ang{20} below the face and with gaze points above the camera. Relative to the
camera, the gaze angle range is \SIrange{0}{40}{\degree} in pitch and
\SI{+-40}{\degree} in yaw. The head pose range is \SIrange{10}{30}{\degree} in
pitch and \SI{+-20}{\degree} in yaw. Specifically, we set the fields in the
tool to $\left[-20,0,10,20\right]$ and $\left[0,0,10,20\right]$. Since we know
the geometry of the eye surface, we have the opportunity to add glints from a
coaxial light source. We generate one million UnityEye images and split
\num{80}/\num{10}/\num{10} for training/validation/test. To reduce compression
artifacts, we generate the images at \num{1024 x 768}~pixels and rescale them
to match the camera of the NIR dataset.

The eyes are placed at \SI{65}{\centi\meter}, with gaze targets randomly placed
at a $z$-depth of one of \SIlist{-30; 0; +30}{\centi\meter}. We sample the
interocular distance from $\mathcal{N}(\SI{63}{\milli\meter},
\SI[parse-numbers=false]{3.5^2}{\milli\meter\squared})$~\cite{Gordon2014}. See
Figure~\ref{fig:unity_roi} for an example image. We use the center of the
eyelid annotations as the eye detections. To make the detections less perfect,
we randomly offset them in a circular disk with a radius equal to
\SI{3}{\percent} of the interocular distance.

\begin{figure}
	\centering
	\input{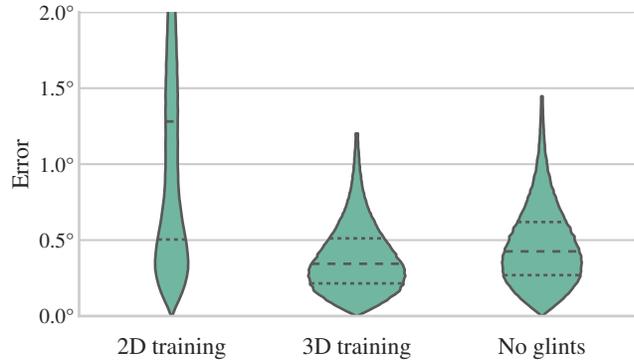}
	\caption{\
		Distribution of angle errors on the UnityEyes dataset.
		\emph{2D~training} has all gaze targets in the same plane. For
		\emph{3D~training}, the gaze targets are distributed among three
		planes. \emph{No~glints} also has the gaze targets in three planes, but
		in this case no glints were added to the images. In all cases, the test
		data has gaze targets in three planes.}\label{fig:unity}
\end{figure}

The results are shown in Figure~\ref{fig:unity}. We see that the 2D training
data result in very large errors. This was also reflected in the distances
between estimated gaze rays and true gaze origins. On this synthetic data, we
see that glints do improve the accuracy by a small factor.

\subsection*{Improving the distance}

\begin{figure}
	\centering
	\input{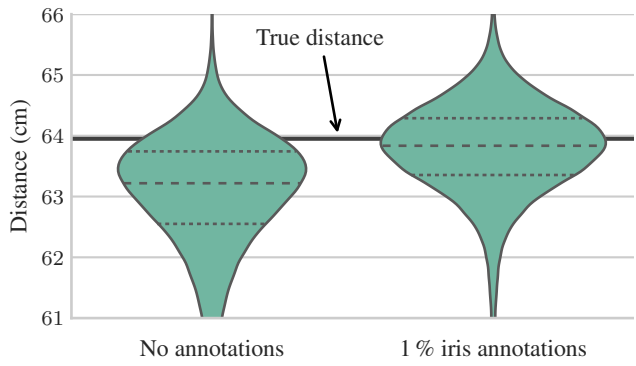}
	\caption{\
		Distribution of estimated distances to the iris on the UnityEyes
		dataset, with and without iris annotations.}\label{fig:distance}
\end{figure}

While the generated gaze rays as such are close to the true gaze origins, there
is nothing forcing the estimated gaze origins to coincide with any physical
feature on the eye, and the origins tend to be ``floating'' along the gaze ray.
In an attempt to improve the estimates, we used iris metadata to annotate a
\SI{1}{\percent} subset of the UnityEyes image with iris centers. We then added
an additional term to the cost function, an $L^2$ loss on the distance between
the iris center and the 2D gaze origin. The distances to the estimated gaze
origins are shown in Figure~\ref{fig:distance}. We see that sparse iris
annotations improves the consistency of the distance estimate.

\end{document}